\documentclass[conference,letterpaper]{ieeetran}
\IEEEoverridecommandlockouts
\usepackage{graphicx} 
\usepackage{siunitx}
\usepackage{multirow}

\usepackage[utf8]{inputenc}

\usepackage{dcolumn}           

\usepackage{amsmath}
\DeclareMathSizes{7}{7}{3}{3} 
\usepackage{pifont}
\usepackage{amsfonts}
\usepackage{amssymb}

\usepackage{float}

\usepackage{titlesec}
\usepackage{booktabs}
\usepackage{csquotes}

\usepackage{subcaption}

\usepackage{secdot}
\usepackage{epstopdf}

\title{\LARGE \bf Model-Based Lookahead Reinforcement Learning for in-hand manipulation}
  \author{\IEEEauthorblockN{Alexandre Filipe Gonçalves Lopes}
 \IEEEauthorblockA{\textit{Instituto Superior T\'ecnico}\\
 Lisboa, Portugal \\
 alexandre.f.g.lopes@tecnico.ulisboa.pt}
 \and
 \IEEEauthorblockN{Catarina Barata, Plinio Moreno}
 \IEEEauthorblockA{\textit{Instituto de Sistemas e Rob\'otica} \\
 \textit{Instituto Superior T\'ecnico}\\
 Lisboa, Portugal \\
 ana.c.fidalgo.barata@tecnico.ulisboa.pt,plinio@isr.tecnico.ulisboa.pt}
 }

\begin{document}
\maketitle


%



\begin{abstract}

In-Hand Manipulation, as many other dexterous tasks, remains a difficult challenge in robotics by combining complex dynamic systems with the capability to control and manoeuvre various objects using its actuators. This work presents the application of  a previously developed hybrid Reinforcement Learning (RL) Framework to In-Hand Manipulation task, verifying that it is capable of improving the performance of the task. The model combines concepts of both Model-Free and Model-Based Reinforcement Learning, by guiding a trained policy with the help of a dynamic model and value-function through trajectory evaluation, as done in Model Predictive Control. This work evaluates the performance of the model by comparing it with the policy that will be guided. To fully explore this, various tests are performed using both fully-actuated and under-actuated simulated robotic hands to manipulate different objects for a given task. The performance of the model will also be tested for generalization tests, by changing the properties of the objects in which both the policy and dynamic model were trained, such as density and size, and additionally by guiding a trained policy in a certain object to perform the same task in a different one. The results of this work show that, given a policy with high average reward and an accurate dynamic model, the hybrid framework improves the performance of in-hand manipulation tasks for most test cases, even when the object properties are changed. However, this improvement comes at the expense of increasing the computational cost, due to the complexity of trajectory evaluation.
\\

\end{abstract}

\section{Introduction}
\label{sec:intro}

In-hand Manipulation is a complex task that requires the ability to precisely and efficiently manipulate different sized and shaped objects. In the context of assembly lines, human workers outperform robots in tasks that perform multiple actions with dexterity, precision and adaptability, usually involving various tools and objects, such as engine assembling in the automobile industry. This difference in efficiency is mainly due to the difference in complexity between a human hand and a robotic gripper.

While very complex robotic hands that could be used as a substitute for the grippers already exist, such as the Shadow Hand\footnote{https://www.shadowrobot.com/products/dexterous-hand/}, using them comes with another challenge: the system needs to learn how to use the hand to manipulate objects. A way to do this is to use a method that takes inspiration from how humans learn, Reinforcement Learning (RL). Although RL explain the main mechanism behind reward learning, execution of real-time tasks by humans requires additional components such as sensorial anticipation, in order to deal with latencies \cite{WOLPERT1998338}. These anticipatory mechanisms rely on internal simulations of the forthcoming sensorial input, which works like the Model Predictive Control (MPC) approach \cite{mpc-1982,RICHALET19931251}. From the RL point of view, MPC corresponds use Model-Based RL (MBRL) in a short-time horizon. Since Model-Based requires an accurate model of the dynamics of the environment, it is very difficult to apply it to in-hand object manipulation. Thus, we follow the hybrid model that combines Model-Free RL (MFRL) and MBRL. One of these methods is Model-Based Lookahead Reinforcement Learning \cite{hong2019model}. The main focus of this work is to apply this method with robotics, and conclude if the promising results achieved in the previous tests are still achievable for in-hand manipulations tasks using a dexterous robotic hand.

Our method requires exploring implementations of both Model-Free Algorithms and learning dynamic models. The main objective of this work is to compare the performance of Model-Based Lookahead RL to the performance of a commonly used Model-Free RL algorithm. To achieve this, the first objectives are to obtain the required models to perform the task through a Model-Free method and obtaining a model of the system dynamics. Hybrid RL methods consider the information obtained from both Model-Free and Model-Based methods, which brings improvements in various scenarios. Our main goal is to obtain both models with high accuracy through RL methods, and combine them to achieve a better performance than a commonly used Model-Free RL algorithm, when tested for an in-hand manipulation task, using different objects and different robotic systems, such as fully and under actuated systems. Then, we analyze the robustness and generalization of the hybrid model when applied to tasks that deviate from the original task it was trained on, and see if it still outperforms a Model-Free RL approach. The final goal is to conclude the computational costs of Model-Based Lookahead RL, in comparison with its MFRL counterpart, such as iterations per second and runtimes.

\section{Related Work}
\label{sec: backg}

This section presents the current state of the art on the topics of in-hand manipulation and RL, discussing the various methods used, its achievements, and how they are relevant to the objectives defined for this work.

\subsection{Introduction of Notation}

RL problems can be described as Markov Decision Processes \cite{sutton2018reinforcement}, using the tuple $\{S,A,R,T,\gamma\}$. Here, $S$ represents the state space, and $A$ the action space. $R(s,a)$ represents a reward function, which returns the immediate reward for taking an action $a$ in a state $s$. $T(s'|s,a)$ is the dynamics of the system, which returns the probability distribution of the next states $s'$ after taking action $a$ in state $s$. Finally $\gamma \in [0,1]$ is a discount factor, applied to the sum of immediate rewards $G$, where $G =\sum_{t=0}^{\infty} \gamma^{t}R(s_t,a_t)$. The objective of RL is to obtain a policy, $\pi$ (i.e. controller in our case) that selects actions in order to maximize the rewards.

\subsection{Model-Free Reinforcement Learning} \label{mfrl}

In cases such as in-hand manipulation of objects, traditional models of the dynamics require to know all contact points and forces. These forces can be computed only in the case of rigid objects, but robotic hands are usually built with elastic components. Thus, the dynamics model $T$ suffer from a large reality gap that limits the application of purely MPC models. MFRL approaches the problem by learning policies from exploration, without having to learn a model of its dynamics. Two ways of approaching MFRL are policy-based and value-based methods: (i) In value function methods, the objective is to learn the value of the states $V$, or the value of state-action pairs $Q$, that is, the expected rewards obtained when following a policy $\pi$, starting at a certain state/state-action pair. The optimal value function $V^*(s)$ is defined as $V^*(s)=\max _\pi \mathbb{E}\left[G_t \mid s_t=s\right]$, and the optimal action-value function $Q^*(s,a)$ is defined as $Q^*(s, a)=\max _\pi \mathbb{E}\left[G_t \mid s_t=s, a_t=a, \pi\right]$. (ii) In policy-based methods the objective is to parameterize a policy. The policy $\pi$ will depend on a parameter vector $\theta$. This approach is based on the policy gradient theorem, which gives a proportional equation to the gradient of the performance measure, $J(\theta)$, updating the parameter vector as $\theta_{t+1} = \theta_t  + \alpha\nabla J(\boldsymbol{\theta})$. The gradient of the performance measure is derived in \cite{sutton2018reinforcement}.

We explore the use of an Actor-Critic method, which uses both a value function (critic) and a policy parametrization  (actor). While the actor decides on which action to take, the critic will evaluate the action-state pair and give the actor a measurement of how good the action is. These algorithms, such as PPO \cite{schulman2017proximal} and SAC \cite{haarnoja2018soft} have had success in in-hand manipulation tasks, such as \cite{andrychowicz2020learning}, where the objective was to manipulate a cube with labeled sides to the correct configuration using vision. 

\subsection{Model-Based Reinforcement Learning} \label{mbrl}

A different approach to MFRL which is more data efficient \cite{deisenroth2013survey} is MBRL. By interacting with the environment, MBRL learns a model which represents the dynamics of the system. After learning, the model is then used for planning. Planning can be done offline, where the algorithm uses the information from the model to obtain a policy, or online, such as with MPC. MPC has shown very good results at selecting the right action while considering a small (i.e. non-infinite) horizon. Its sucess comes from very good model of the dynamics of the environment. While MBRL learning models require less data then MFRL, they require a very accurate dynamics model. This dynamics model can be modeled as deterministic or probabilistic. Deterministic models assume that the environment is predictable and without randomness or noise, meaning that for the same state and action the output will always be the same, $s' = f(s,a)$, where as probabilistic models output the probability distribution of the system dynamics $T(s'|s,a)$. 
In robotic tasks, success has been found in using deep NNs for the dynamics model. In \cite{lenz2015deepmpc}, a robot using a knife is used to cut different foods. The NN model is then applied online with the use of an MPC controller.

\subsection{Combining MFRL and MBRL} \label{mbrl}
While current efforts study how to develop a generalist world model (i.e. a multiple task that represents the dynamics as latent representation) that combines MFRL and MBRL, these methods require very large computational resources \cite{Hansen2022tdmpc,hansen2024tdmpc2}. We focus on previous works that (i) are task specific and develop models with (ii) relatively small number of parameters \cite{lowrey2018plan,bhardwaj2021blending,bejjani2021learning,hong2019model}. These works leverage the strengths of MFRL and MBRL in a hybrid manner, such as the Model-based Lookahead Reinforcement Learning \cite{hong2019model}. We select \cite{hong2019model} because the action selection approach is not as greedy as in \cite{lowrey2018plan}, and has less parameters to tune than in \cite{bhardwaj2021blending,bejjani2021learning}. In \cite{hong2019model}, the authors propose that, after training a policy and value-function through an Actor-Critic Method, and a dynamic model with the extracted datasets from training, trajectory evaluation is then applied during online planning, as done with an MPC. This approach will serve as the base method for this work. The model is divided in two parts: the training phase, and the evaluation phase:
The training phase begins by training in parallel with the same data, the policy $\pi_{\theta^\pi}$, the value function $V_{\theta^V}$ and the transition model $f_{\theta^f}$. All of these are trained as NNs, as parametric functions with parameters $\theta^\pi$,$\theta^V$,and $\theta^f$ respectively.\\
To improve the dynamics model $f_{\theta^f}$, the data will be obtained from an exploratory policy $\pi'$. From these a set of states $\tau'$ is obtained, defined over a period $T$: $\tau_{\pi^{\prime}}=\left[\boldsymbol{s}_1, a_1, \boldsymbol{s}_2, \cdots, \boldsymbol{s}_{T+1}\right]$. The trajectories are then truncated into various entries containing $\{s_t,a_t,s_{t+1}\}$ and stored in a dataset $\mathcal{D}$.\\
With the dataset collected, the dynamics model is trained using gradient descent, and the policy and value functions using a policy-based and \linebreak value-based methods, respectively, or an actor-critic method. 

During the evaluation phase, the approach uses online planning through an MPC. To discuss this further, the evaluation can be explained in three steps: Trajectory Sampling; Trajectory Evaluation; Action Selection. During trajectory sampling, the trajectories are obtained from the use of both the learned policy and dynamics model as such:
\begin{equation}
\label{trajectory}
\hat{s}_{1}^{n}\,=\,s_{t};\;\hat{s}_{h+1}^{n}\,=\,f_{\theta^f}\,(\hat{s}_{h}^{n},\hat{a}_{h}^{n});\;\hat{a}_{h}^{n}\,\sim\pi_{\theta^{n}}(a|\hat{s}_{h}^{n}).
\end{equation}
The trajectory actions are obtained from the \linebreak learned policy $\pi_{\theta^\pi}$, and with the state-action pair the next state from $f_{\theta^f}$ can be obtained, over $H$ steps. It is important to mention that in this work, the learned dynamics model $f_{\theta^f}$ is deterministic, which assumes the same output state vector $\hat{s}_{h+1}^{n}$ given the same input pair $(\hat{s}_{h}^{n},\hat{a}_{h}^{n})$. Our dynamic models are deterministic as well.

From the current state, $n$ of these trajectories are computed, and proceed to associate a reward to each one. This is done in the Trajectory Evaluation step: the function to maximize utilizes the global information from the learned value-function $V_{\theta^V}$, where the total discounted reward of the trajectory is given as:
\begin{equation}
\label{reward}
\hat{G}(\hat{s}_{1:H+1}^{n},\hat{a}_{1:H}^{n})=\sum_{h=1}^{H}\gamma^{h-1}R(\hat{s}_{h}^{n},\hat{a}_{h}^{n})+\gamma^{H}V_{\theta^{V}}(\hat{s}_{H}^{n}).
\end{equation}
After evaluation, the trajectory with the highest reward could be chosen, and its first action would be used, but this might be too optimistic \cite{hong2019model}, as the simulated rewards are obtained using a learned dynamics model and policy, and not an actual model of the environment. A way to reduce the approximation errors from this method is to, instead of greedily choosing the best trajectory, using the average of the best $E$ trajectories, where $E$ is a fixed chosen value. This is performed by using a sorting algorithm, and finally outputting the averaged first action of the best trajectories.

Following the description of the Model-based Lookahead RL framework, its implementation in real robotics comes with some limitations:

Unlike most approaches in which there is only one NN to train, this method needs to train three separate NNs. Even if the model has a high sample efficiency reducing the size of the needed data set, the computational resources (such as number of GPU tensor cores and available memory) needed to train the networks in parallel is much higher, especially in highly complex problems such as in-hand manipulation.
Another issue is the high amount of hyperparameters. Fine tuning these values will be time consuming, and will require more experimentation. Finally, as mentioned in \cite{hong2019model}, the results obtained from using horizon values 5 and 20 are very similar. As the trajectories are obtained from an approximated dynamics model, they will accumulate errors, a problem that may result in poor prediction of trajectories for very large horizons. This may hinder performance, as longer horizons allow for more extensive exploration \cite{lowrey2018plan}.
\section{Methodology}
\label{sec: Methodology}

On in-hand manipulation, many tasks require access to the position or orientation of an object (Ex: rotating a cube to a labeled face \cite{andrychowicz2020learning}). With the current information the dexterous hand provides, this is not possible, as it only provides access to the feature values of the hand and its contact points. A way to obtain the pose of the objects, is to use a pose detector, which can be substituted in most physic simulators. Because of this, it is a design choice to fully develop this work in a simulated environment. 

\subsection{Dynamics}

The simulator used in this work is Nvidia Isaac Gym \cite{liang2018gpu}. This simulator provides a physics engine, as well as integration with various frameworks such as SKRL \cite{serrano2023skrl}, a python library for RL which is integrated with Nvidia Isaac Gym, the PyTorch framework \cite{NEURIPS2019_9015}, and Isaac Gym Benchmark Environments (IsaacGymEnvs) \cite{makoviychuk2021isaac}, a repository containing various tasks, objects and other assets. 
Another important feature of Isaac Gym is the use of multi-environments, which will increase the number of training samples. The models will be trained with $2^{13}$ parallel environments. The simulator gives access to the state vectors of the simulated objects over each timestep, such as position, orientation, velocity and acceleration, where the position, linear velocity and linear acceleration are given as a three dimensional vector $(x,y,z)$ and the orientation is defined as quaternions $(q_1,q_2,q_3,q_4)$, removing the need for a pose detector.

\begin{figure}
    \centering
    \includegraphics[width=0.367\columnwidth]{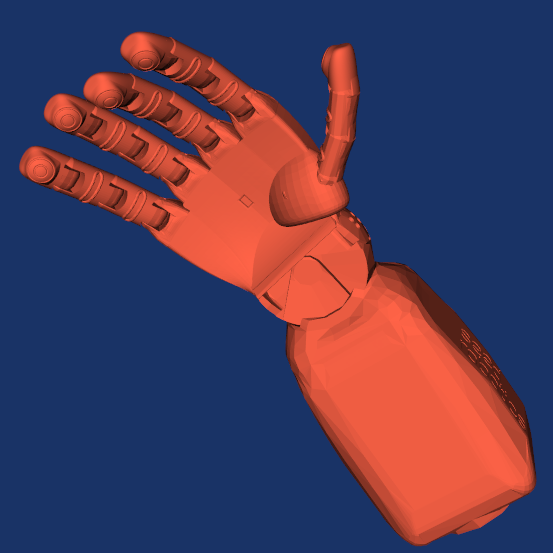}\includegraphics[width=0.5\columnwidth]{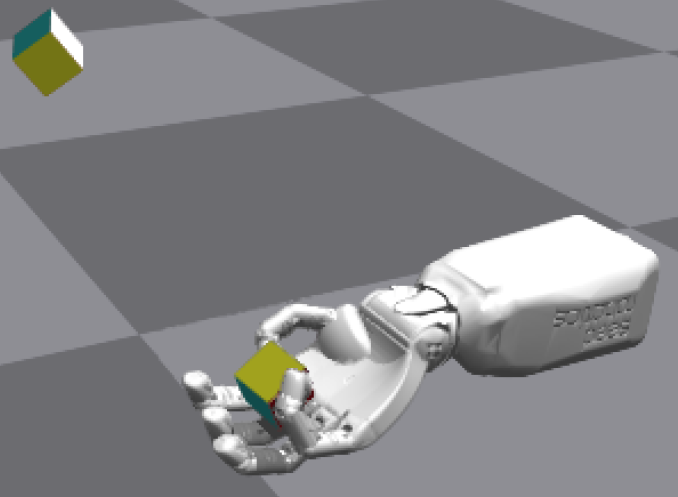}
    \caption{Left-hand side: Render of the RH8DR hand. The real world hand is underactuated, having one string per finger. We consider two hand options: (i) Fully actuated and (ii) underactuated. Rigth-hand side: in-hand manipulation of a cube in simulation.}
    \label{fig:enter-label}
\end{figure}

As for the robotic hand, this work will use the provided URDF models by Seed Robotics for the RH8DR robotic hand\footnote{https://www.seedrobotics.com/rh8d-adult-robot-hand}, displayed in Figure \ref{fig:enter-label}. 
We study two different test cases for the actuation and control of the simulated robotic hand: (i) fully-actuated and (ii) under-actuated.

In the fully actuated version, each joint will be actuated and controlled individually. In Isaac Gym, each of the DoF will use a PD controller, using position as the reference value $u(t)$. The proportional component of the error $e(t)$ will be the position error and the derivative component will be the velocity error. Since all joints in the model are revolute, these values are in $rad$ and $rad/s$ respectively.

Looking at the state vector, each DoF provides its position ($rad$) and velocity ($rad/s$), 
giving 2 variables per joint. As for the rigid bodies (object), the API from Isaac Gym obtains their position $(x,y,z)$ $[m]$,  orientation $(q_1,q_2,q_3,q_4)$, and linear velocity $(x,y,z)$ $[m/s]$, totaling 10 variables for the object. Finally the state vector will also include the previous actions used. While there are 19 DoF available, only 17 are used, as for the palm only pitch is used, meaning yaw and roll joints are fixed, since the following manipulation tasks do not require them as they only hinder performance. This gives a total state-vector containing 61 variables. As for the action space, there are 17 variables, one for each DoF used.

As for the underactuated hand, we mimic the actuation of the real RH8DR, with three actuators for the wrist, two for the thumb, one for the index finger, one for the middle finger, and a single actuator for both the ring and little finger. These actuators need to replicate the tendons used to control the finger joints, as to obtain the synergies of the real robot.\\
To perform this in Nvidia Isaac Gym, this work uses a system based on the work \cite{catalano2014adaptive} which, given the position and linear velocity of the finger joints, as well as the action variables, can calculate based on a spring coefficient matrix $K$ and Synergy Coefficient matrix $S$ the torque value of those finger joints. Since all of the DoF are revolute, the inputs are torque values in $Nm$. The rest of the joints are all controlled in position, which are the wrist joints and thumb-palm joint, in radians.

Looking at the state vector, the simulation still has access to both position ($rad$) and velocity \linebreak ($rad/s$) of each joint, even if they are not independently actuated, meaning there are the same variables as the fully-actuated system, except for the number of previous actions, which are now 6 actuators. This results in a vector state with 50 variables.


\subsection{Manipulation Task}

We focus on consecutively rotating an object on the palm of the hand to goal orientations, for a fixed period of time. This task is obtained from the Isaac Gym Benchmark Environments, which is based on the task \cite{andrychowicz2020learning} from OpenAI. The task is obtained from utilizing two instances of the same object: object, and goal object. For a defined maximum episode time, the object is spawned on top of the palm of the hand on a random orientation, where the goal is to be inside a small range off the orientation of the goal object, which is spawned in the simulation at the same time, with a random orientation as well. If the object matches the orientation within the defined maximum episode time (600 timesteps), the goal object is now re-spawned with a new orientation, while the object continues in the same position, and must match the new orientation. The reset condition will be if the object has fallen from the palm, represented by a defined fall distance parameter. The number of consecutive times the object matches the orientation of the goal object without failing will be recorded, as well as the average episode time. The task will be performed for three different objects: a cube, an egg, and a parallelepiped. The three objects have varying geometrical complexities which will result in tasks with different difficulties, which may give more insight on the performance gain of Model-Based Lookahead RL.
\section{Experiments and Results}
\label{sec:resul}


\subsection{Obtaining the Policy} \label{sec:policy}

\begin{figure}[h]
    \centering
  \subcaptionbox{Fully-actuated system}{
    \includegraphics[scale=0.5]{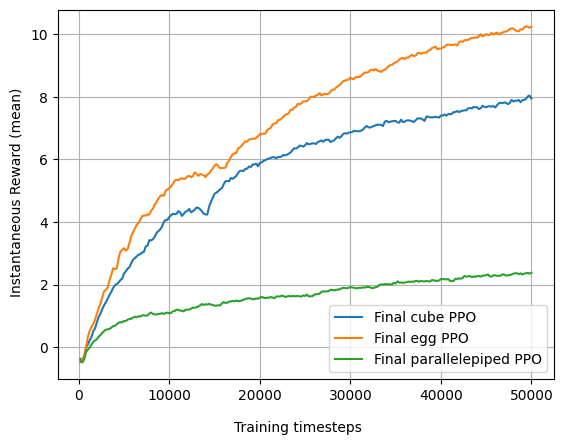}}
 \hspace{0.4cm}
   \subcaptionbox{Under-actuated system}{
    \includegraphics[scale=0.5]{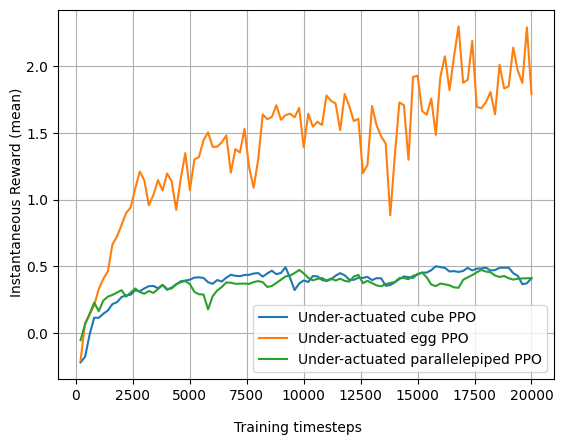}}

\vspace{0.5cm}
\caption{Final training Average Reward for the cube, egg and parallelepiped}
\label{fig:two}
\end{figure}

The first step of the experimental plan consists in obtaining a well performing policy, that can achieve the defined goals of the task from section \ref{sec: Methodology}. The training will utilize the actor-critic PPO algorithm, with $2^{13}$ parallel environments, over 500000 time steps each, totaling 409,6 million samples. The best results obtained for each object utilize a dense Feed Forward Neural Network, with a common initial network, followed by a separation into two distinct networks: one for the mean $\mu$, and the second one for the standard deviation $\sigma$. The reward function will use rotation difference between object and target object, a reach goal bonus, a small fall penalty, L2 regularization over the action space to penalize large actions around the maximum position of the joints; and action smoothing, which penalizes large differences between the current position and target position. This results in the following graph \ref{fig:two}, which gives the Average Reward of each policy.

\begin{table*}[t]
\centering
\caption{Final model using fully-actuated system: Additional results for the cube, egg and parallelepiped}
\label{tab:train_final}
\begin{tabular}{|c|c|c|c|}
\hline
\textbf{Object} & \textbf{Average Reward} & \textbf{Consecutive Sucesses} & \textbf{Average Episode Length} \\ \hline
Cube PPO           & 8.279$\pm$ 0.139 & 18.1 & 539.6 \\ \hline
Egg PPO            & 10.998$\pm$ 0.110 & 24.1 & 535.6 \\ \hline
parallelepiped PPO & 2.461$\pm$ 0.046  & 4.3  & 432.1 \\ \hline
\end{tabular}%
\end{table*}

\begin{table*}[t]
\centering
\caption{Final Model using under-actuated system: Additional results for the cube, egg and parallelepiped}
\label{tab:train_tendon}
\begin{tabular}{|c|c|c|c|}
\hline

\textbf{Object} & \textbf{Average Reward} & \textbf{Consecutive Sucesses} & \textbf{Average Episode Length} \\ \hline
Cube PPO           & 0.449$\pm$0.031 & 0.1 & 521.5 \\ \hline
Egg PPO            & 1.654$\pm$0.159 & 1.1 & 392.2 \\ \hline
parallelepiped PPO & 0.501$\pm$0.054 & 0.0 & 498.4 \\ \hline
\end{tabular}%
\end{table*}

From the results obtained, it is clear that the fully-actuated system outperforms the under-actuated system for all objects. The under-actuated results are shown until 20k instead of 50k timesteps, as the policies all converged around this point. In comparison with the fully-actuated system, the final Average Reward of all objects is much smaller, only obtaining a value above 1 consecutive success in the egg test, with the cube and parallelepiped obtaining close to or 0 successes. By limiting the joint movement of all fingers, especially by linking the small and ring finger, the state space is much more constrained than the fully-actuated system, making complex movements such as rotations of objects much harder. 

An interesting result is the better performance of the egg compared to the cube, as the egg has a much more complex geometry which leads to more complex interactions between the hand and object. A possibility for this result may come from the fact that the policy, in contrast to results such as \cite{andrychowicz2020learning}, during manipulation the robotic hand tends to roll the object on the palm, assisted by the finger to rotate them, instead of picking up the object with opposing fingers, holding it with the thumb. In the obtained results of this work, the thumb is barely used, only applying forces on the object without much result in movement. By lowering the pitch DoF of the palm, the robot rotates the object with the pinky, ring, middle and index fingers on the palm instead of picking it up, which leads to better results when the objects have rounder surfaces, such as the egg, as opposed to the cube and parallelepiped. Since the cube has a smaller and uniform size, even with sharp edges, the robot still manages to rotate it with ease, however, the addition of different sizes for the x axis of the parallelepiped (the x axis is double the value of y and z, which are the same size as the cube) results in harder movements for the robot to perform, reducing the overall performance significantly.

Unlike the fully-actuated system however, in the under-actuated case the cube does not outperform the parallelepid, having both similar performance metrics, meaning that this system struggles with objects that have straight sides and sharp edges. This problem although present in the fully-actuated system as the egg had the best performance, is much more noticeable in this system, as the cube also has a poor performance.

\subsection{Obtained the Dynamics Model} \label{sec:dynmodel}

Having the extracted the datasets from each task, it is now possible to train the dynamic models of each one. The dynamic models with highest performance metrics obtained utilize the finger joints and objects position and linear velocity. Since position control utilizes both position and velocity errors in the PD controller, these features are essential for obtaining low prediction errors for the finger joints. For training the neural network, Supervised Learning is performed, aided with the Adam optimizer \cite{kingma2014adam}. The configuration used repeats a block containing a Linear Fully Connected Layer, followed by Batch Normalization, a ReLU activation function and a dropout layer with probability of 20\%. This is enough for the under-actuated system, however the for the fully-actuated system due to the large action-space, a modular approach is added, using an additional network (Single Joint Model) at the start which takes as input the joint position, velocity, and action, and outputs the predicted position and velocity. This is done for each joint, as each joint is independent from each other, simplifying the prediction of the joint values, which are then used as input in the Dynamic Model. The results of these network are shown in tables \ref{tab:dynmodel_modular} and \ref{tab:dynmodel_tendon}:

\begin{table*}[t]
\centering
\caption{Modular Dynamic Model for fully-actuated system: Losses in RMSE and MAE for the train and test sets}

\label{tab:dynmodel_modular}
\resizebox{\textwidth}{!}{%
\begin{tabular}{|c|lc|clccl|}
\hline
\textbf{} &
  \multicolumn{2}{c|}{\textbf{RMSE}} &
  \multicolumn{5}{c|}{\textbf{MAE}} \\ \hline
\textbf{Object} &
  \multicolumn{1}{c|}{\textbf{\begin{tabular}[c]{@{}c@{}}Overall Error\\ Train set\end{tabular}}} &
  \textbf{\begin{tabular}[c]{@{}c@{}}Overall Error\\ Test set\end{tabular}} &
  \multicolumn{1}{c|}{\textbf{\begin{tabular}[c]{@{}c@{}}Joint Position \\ Error (rad)\end{tabular}}} &
  \multicolumn{1}{c|}{\textbf{\begin{tabular}[c]{@{}c@{}}Joint Velocity \\ Error (rad/s)\end{tabular}}} &
  \multicolumn{1}{c|}{\textbf{\begin{tabular}[c]{@{}c@{}}Object Position \\ Error (m)\end{tabular}}} &
  \multicolumn{1}{c|}{\textbf{\begin{tabular}[c]{@{}c@{}}Object Orientation \\ Error\end{tabular}}} &
  \textbf{\begin{tabular}[c]{@{}l@{}}Object Velocity \\ Error (m/s)\end{tabular}} \\ \hline
\textbf{Cube} &
  \multicolumn{1}{l|}{0.1290} &
  0.1415 &
  \multicolumn{1}{c|}{0.0529} &
  \multicolumn{1}{l|}{0.1211} &
  \multicolumn{1}{c|}{0.0189} &
  \multicolumn{1}{c|}{0.0694} &
  0.0922 \\ \hline
\textbf{Egg} &
  \multicolumn{1}{l|}{0.1573} &
  0.1645 &
  \multicolumn{1}{c|}{0.0438} &
  \multicolumn{1}{l|}{0.1482} &
  \multicolumn{1}{c|}{0.0083} &
  \multicolumn{1}{c|}{0.0657} &
  0.1179 \\ \hline
\textbf{Parallalepiped} &
  \multicolumn{1}{l|}{0.1160} &
  0.1301 &
  \multicolumn{1}{c|}{0.0520} &
  \multicolumn{1}{l|}{0.1137} &
  \multicolumn{1}{c|}{0.0151} &
  \multicolumn{1}{c|}{0.0696} &
  0.0999 \\ \hline
\end{tabular}%
}
\end{table*}

\begin{table*}[t]
\centering
\caption{Dynamic Model for the under-actuated system: Losses in RMSE and MAE for the train and test sets}
\label{tab:dynmodel_tendon}
\resizebox{\textwidth}{!}{%
\begin{tabular}{|c|lc|clccl|}
\hline
\textbf{} &
  \multicolumn{2}{c|}{\textbf{RMSE}} &
  \multicolumn{5}{c|}{\textbf{MAE}} \\ \hline
\textbf{Object} &
  \multicolumn{1}{c|}{\textbf{\begin{tabular}[c]{@{}c@{}}Overall Error\\ Train set\end{tabular}}} &
  \textbf{\begin{tabular}[c]{@{}c@{}}Overall Error\\ Test set\end{tabular}} &
  \multicolumn{1}{c|}{\textbf{\begin{tabular}[c]{@{}c@{}}Joint Position \\ Error (rad)\end{tabular}}} &
  \multicolumn{1}{c|}{\textbf{\begin{tabular}[c]{@{}c@{}}Joint Velocity \\ Error (rad/s)\end{tabular}}} &
  \multicolumn{1}{c|}{\textbf{\begin{tabular}[c]{@{}c@{}}Object Position \\ Error (m)\end{tabular}}} &
  \multicolumn{1}{c|}{\textbf{\begin{tabular}[c]{@{}c@{}}Object Orientation \\ Error\end{tabular}}} &
  \textbf{\begin{tabular}[c]{@{}l@{}}Object Velocity \\ Error (m/s)\end{tabular}} \\ \hline
\textbf{Cube} &
  \multicolumn{1}{l|}{0.0899} &
  0.1002 &
  \multicolumn{1}{c|}{0.0346} &
  \multicolumn{1}{l|}{0.0813} &
  \multicolumn{1}{c|}{0.0125} &
  \multicolumn{1}{c|}{0.0456} &
  0.0677 \\ \hline
\textbf{Egg} &
  \multicolumn{1}{l|}{0.1062} &
  0.1099 &
  \multicolumn{1}{c|}{0.0460} &
  \multicolumn{1}{l|}{0.0930} &
  \multicolumn{1}{c|}{0.0137} &
  \multicolumn{1}{c|}{0.0444} &
  0.0675 \\ \hline
\textbf{Parallalepiped} &
  \multicolumn{1}{l|}{0.0840} &
  0.0855 &
  \multicolumn{1}{c|}{0.0353} &
  \multicolumn{1}{l|}{0.0664} &
  \multicolumn{1}{c|}{0.0090} &
  \multicolumn{1}{c|}{0.0372} &
  0.0459 \\ \hline
\end{tabular}%
}
\end{table*}

\subsection{Model-based Lookahead Reinforcement Learning}

Having obtained the policy and value-function from training a PPO algorithm, and training a dynamic model from the datasets extracted for all objects with high reward/minimal error, it is now possible to combine them in evaluation mode as explained in section \ref{mbrl}. As trajectory evaluation is referred as MFRL-MPC \cite{hong2019model}, for further tests the model using PPO as base for the policy will be referred as PPO-MPC.

The first case is to test the results using the correct policy and dynamic model for the corresponding task. For example, for the task of rotating a cube, utilizing the policy and dynamic model trained on that same cube. The policies used will be at the maximum previous timestep of 50k for each object. The model will use an horizon of 2 (hyperparameter $H$), 1024 trajectories (hyperparameter $N$), and average over the 2 best trajectories (hyperparameter $E$). This test results in the following table:

\begin{table*}[t]
\centering
\caption{Comparison between running the policy trained with PPO and trajectory evaluation with PPO-MPC for all objects}
\label{tab:result_best}
\begin{tabular}{|c|c|c|c|}
\hline
\textbf{Object} & \textbf{Average Reward} & \textbf{Consecutive Sucesses} & \textbf{Average Episode Length} \\ \hline
Cube PPO-MPC           & 9.344$\pm$0.104  & 20.4 & 539.2 \\ \hline
Cube PPO               & 8.2787$\pm$0.139 & 18.1 & 539.6 \\ \hline
Egg PPO-MPC            & 12.097$\pm$0.136 & 26.6 & 547.6 \\ \hline
Egg PPO                & 10.998$\pm$0.110 & 24.1 & 535.6 \\ \hline
parallelepiped PPO-MPC & 2.963$\pm$0.073  & 5.3  & 477.9 \\ \hline
parallelepiped PPO     & 2.461$\pm$0.046  & 4.3  & 432.1 \\ \hline
\end{tabular}%
\end{table*}

Looking at the results from table \ref{tab:result_best}, there is a significant increase in both Average Reward and Consecutive Successes in all cases, and a small increase in terms of Average Episode Length for the egg and parallelepiped. Since the maximum episode length is defined as 600, which given a clock of 60Hz has a runtime of 10s, additional metrics can be studied:

\begin{itemize}
    \item For the cube, there is Reward increase of 1.065, a consecutive success increase of 2.2, and the PPO achieves the goal on average over 33.150 timesteps while the PPO-MPC achieves success over 29.557 timesteps. 
    \item For the egg, there is Reward increase of 1.099, a consecutive success increase of 2.5, and the PPO achieves the goal on average over 24.896 timesteps while the PPO-MPC achieves success over 22.556 timesteps. 
    \item For the parallelepiped, there is Reward increase of 0.502, a consecutive success increase of 1.0, and the PPO achieves the goal on average over 139.535 timesteps while the PPO-MPC achieves success over 113.208 \linebreak timesteps. 
\end{itemize}

\subsection{Generalization Test} \label{generalize}

\begin{table*}[h]
\centering
\caption{Generalization tests: changing weights for the cube task}
\label{tab:weight_cube}
\begin{tabular}{|c|cc|cc|cc|}
\hline
\multirow{2}{*}{\textbf{\begin{tabular}[c]{@{}c@{}}Density\\ Multiplier\end{tabular}}} &
  \multicolumn{2}{c|}{\textbf{Average Reward}} &
  \multicolumn{2}{c|}{\textbf{Average Consecutive Sucesses}} &
  \multicolumn{2}{c|}{\textbf{Average Episode Length}} \\ \cline{2-7} 
 &
  \multicolumn{1}{c|}{\textbf{PPO-MPC}} &
  \textbf{PPO} &
  \multicolumn{1}{c|}{\textbf{PPO-MPC}} &
  \textbf{PPO} &
  \multicolumn{1}{c|}{\textbf{PPO-MPC}} &
  \textbf{PPO} \\ \hline
\textbf{0.5} & \multicolumn{1}{c|}{8.644$\pm$0.143} & 7.459$\pm$0.103 & \multicolumn{1}{c|}{18.3} & 15.9 & \multicolumn{1}{c|}{516.9} & 507.9 \\ \hline
\textbf{1}   & \multicolumn{1}{c|}{9.344$\pm$0.104} & 8.279$\pm$0.139 & \multicolumn{1}{c|}{20.4} & 18.1 & \multicolumn{1}{c|}{539.2} & 539.6 \\ \hline
\textbf{2}   & \multicolumn{1}{c|}{9.734$\pm$0.168} & 8.615$\pm$0.117 & \multicolumn{1}{c|}{21.6} & 19.2 & \multicolumn{1}{c|}{524.3} & 513.6 \\ \hline
\textbf{4}   & \multicolumn{1}{c|}{9.545$\pm$0.169} & 8.563$\pm$0.191 & \multicolumn{1}{c|}{21.7} & 19.5 & \multicolumn{1}{c|}{543.8} & 517.7 \\ \hline
\textbf{8}   & \multicolumn{1}{c|}{8.760$\pm$0.162} & 7.969$\pm$0.204 & \multicolumn{1}{c|}{20.1} & 18.4 & \multicolumn{1}{c|}{550.9} & 549.8 \\ \hline
\textbf{16}  & \multicolumn{1}{c|}{6.637$\pm$0.181} & 6.177$\pm$0.166 & \multicolumn{1}{c|}{15.2} & 14.4 & \multicolumn{1}{c|}{559.2} & 438.5 \\ \hline
\textbf{32}  & \multicolumn{1}{c|}{2.118$\pm$0.096} & 2.105$\pm$0.064 & \multicolumn{1}{c|}{5.5}  & 5.5  & \multicolumn{1}{c|}{537.6} & 524   \\ \hline
\end{tabular}%
\end{table*}
To test the improvements of trajectory evaluation when generalization to different environments, the model will now be utilized to manipulate the cube with varying weights, making $n$ tests with densities multiplied by $2^n$, with $n$ varying from -1 to 4. This is done by changing the object density $\rho$ property in Nvidia Isaac Gym. Table \ref{tab:weight_cube} show the results: 
 The policy trained with PPO manages to generalize well for all cases, starting to decrease the Average Reward and Consecutive Successes from the density $8\rho$ and above. Interestingly, for the densities 2$\rho$ and 4$\rho$, while being different from the original object the policy was trained on, the performance metrics are better than the original, with both higher Average Reward and Consecutive Successes, only being worse in Average Episode \linebreak Lengths. By contrast, in density 0.5$\rho$, all metrics are affected significantly, especially the Consecutive Successes, dropping from 18.1 to 15.9. 
As for the PPO-MPC results, even when using the dynamic model trained of the original cube, there is still improvement in most cases, expect around large differences such as density 32$\rho$. For the smaller increases in density, 2$\rho$ and 4$\rho$, the improvement in performance from PPO-MPC in comparison with the PPO results is large, achieving a similar increase to the original test case. However, for the more extreme cases such as density 32$\rho$, the dynamics are too different to exist any gain in performance by applying trajectory evaluation. However, even if the dynamics model deviates from reality, since the actions are still sampled over the policies action probability distribution, the performance metrics do not fall bellow the values obtained from the PPO test.
In this specific task, even if the objects change, some part of the dynamics still remain equal between them, such as the prediction of the joint position and velocity, which had a low testing error in section \ref{sec:dynmodel} .

\subsection{Guiding a policy on a different task with Trajectory Evaluation}

In this test the objective is to manipulate the egg and parallelepiped using the policy trained on the cube, aided by the corresponding dynamic model. The objective is to see if guiding a policy through trajectory evaluation can substitute the need to train another policy for another task, in this case for a different object. The results are shown in table \ref{tab:guide}:

\begin{table}[h]
\centering
\caption{Guiding a policy trained on the cube, for the egg and parallelepiped tasks, using its corresponding dynamic model}
\label{tab:guide}
\begin{tabular}{|c|cl|cl|cl|}
\hline
\textbf{Model} &
  \multicolumn{2}{c|}{\textbf{Average Reward}} &
  \multicolumn{2}{c|}{\textbf{\begin{tabular}[c]{@{}c@{}}Average\\ Consecutive \\ Sucesses\end{tabular}}} &
  \multicolumn{2}{c|}{\textbf{\begin{tabular}[c]{@{}c@{}}Average\\ Episode \\ Length\end{tabular}}} \\ \hline
Parallelepiped PPO-MPC & \multicolumn{2}{c|}{-0.3099$\pm$0.032} & \multicolumn{2}{c|}{0.5} & \multicolumn{2}{c|}{419.7} \\ \hline
Parallelepiped PPO     & \multicolumn{2}{c|}{-0.441$\pm$0.057}  & \multicolumn{2}{c|}{0.4} & \multicolumn{2}{c|}{397.8} \\ \hline
Egg PPO-MPC            & \multicolumn{2}{c|}{-0.830$\pm$0.025}  & \multicolumn{2}{c|}{0.0} & \multicolumn{2}{c|}{396.3} \\ \hline
Egg PPO                & \multicolumn{2}{c|}{-0.955$\pm$0.014}  & \multicolumn{2}{c|}{0.0} & \multicolumn{2}{c|}{339.5} \\ \hline
\end{tabular}%
\end{table}

As seen in the results from table \ref{tab:guide}, both PPO and PPO-MPC fail to achieve an average of 1 success over the maximum episode length of 600 \linebreak timesteps. In general, by performing the task with the parallelepiped and egg over the policy trained with the cube, unlike the results of section \ref{generalize}, the performance metrics are very poor, as the system is too different for the policy to generalize, which results in an Average Reward of -0.441 and -0.955 respectively for the PPO tests.\\
As for the results of using PPO-MPC, while using trajectory evaluation to guide the policy does increase the performance metrics, it is a very small improvement without noticeable different in Consecutive Successes, a small increase in Average Reward, but a significant difference in Average \linebreak Episode Length, especially the Egg case, going from 339.5 to 396.3 timesteps without falling.

\subsection{Under-Actuated System}

This section will test the Model-Based Lookahead RL Algorithm for the under-actuated system, using the policy obtained in section \ref{sec:policy} and dynamic model in section \ref{sec:dynmodel}. The test will use the same maximum episode length of 600. The results are shown in table \ref{tab:underactuated}.

\begin{table*}[h]
\centering
\caption{Under-Actuated System: Comparison between running the policy trained with PPO and trajectory evaluation with PPO-MPC for all objects}
\label{tab:underactuated}
\begin{tabular}{|c|c|c|c|}
\hline
\textbf{Object} & \textbf{Average Reward} & \textbf{Consecutive Sucesses} & \textbf{Average Episode Length} \\ \hline
Cube PPO-MPC           & 0.487$\pm$0.054 & 0.1 & 527.3 \\ \hline
Cube PPO               & 0.449$\pm$0.031 & 0.1 & 521.5 \\ \hline
Egg PPO-MPC            & 2.001$\pm$0.183 & 1.7 & 461.4 \\ \hline
Egg PPO                & 1.654$\pm$0.159 & 1.1 & 392.2 \\ \hline
parallelepiped PPO-MPC & 0.557$\pm$0.033 & 0.0 & 528.6 \\ \hline
parallelepiped PPO     & 0.501$\pm$0.054 & 0.0 & 498.4 \\ \hline
\end{tabular}%
\end{table*}

Unlike the results obtained in the fully-actuated system, in the under-actuated system the improvement gained from trajectory evaluation is almost negligible in the cube and parallelepiped tasks, but noticeable in the egg task, with an increase of 0.6 Consecutive successes.
Overall, besides the egg task, the results of both cube and parallelepied fail to obtain any significant improvement, even though the obtained dynamic models have similar losses to the dynamic models obtained in the fully-actuated system. Due to the restrictions of the joints, the policies obtained have much smaller Average Rewards and number of Consecutive Successes, as turning the objects to a precise orientation becomes much more difficult with limited movements. Similar to the results of \ref{tab:result_best}, with the Reward Function that uses differences in orientation and action penalty smoothing used for training, policies with higher Average Reward values seem to obtain the largest improvement from trajectory evaluation, which is also similar to overall results obtain in \cite{hong2019model}. As such, when it is not possible to obtain a good policy for a system, the performance gain from Model-Based Lookahead RL is minimal, even if the obtained dynamic model has high accuracy. Since this under-actuated simulation aims to mimic the real-world RH8DR hand, it is likely that in the real system the benefits of trajectory evaluation are also minimal.

\subsection{Computational Cost}

To test the drawbacks of the increased computational cost of trajectory evaluation, both the policy trained with PPO and PPO-MPC (with different combinations of hyperparameters) were tested with only one environment, as to obtain the iterations/second of each model. With the clock values calculated and number of consecutive successes of PPO-MPC and PPO, it is now possible to obtain the runtime needed to achieve success, which results in table \ref{tab:runtime-success}.

\begin{table}[h]
\centering
\caption{Comparison between number of timesteps needed and runtime needed to achieve success, for both running the trained policy from PPO}
\label{tab:runtime-success}
\begin{tabular}{|c|c|c|}
\hline
\textbf{Object}        & \textbf{Timesteps/Success} & \textbf{Runtime/Success (s)} \\ \hline
Cube PPO-MPC           & 29.557                     & 0.857                        \\ \hline
Cube PPO               & 33.149                     & 0.623                        \\ \hline
Egg PPO-MPC            & 22.556                     & 0.654                        \\ \hline
Egg PPO                & 24.896                     & 0.468                        \\ \hline
parallelepiped PPO-MPC & 113.208                    & 3.283                        \\ \hline
parallelepiped PPO     & 139.535                    & 2.623                        \\ \hline
\end{tabular}%
\end{table}

In all results, the number of timesteps with PPO-MPC is smaller than just runnig the policy trained with PPO for each success. However, the trajectory evaluation requires additional time that increases the total Runtime in seconds. It is also important to mention that these values are dependent on the hardware they are ran, as well as the optimization of the algorithm. 

\section{Conclusions}
\label{sec:concl}

\par The focus of this work is to understand the impact of Model-Based Lookahead when applied to in-hand manipulation tasks using two different configurations, a fully-actuated robotic hand and an under-actuated hand. By obtaining a policy and value function by a MFRL method, as well as a dynamic model from the extracted training dataset, the main goal of this work is to verify if performing trajectory evaluation has better performance than only utilizing a policy in various test cases.\\
The challenge of this work stems from the dependency of obtaining an accurate dynamic model, a hard objective due to the complexity of the dynamics of in-hand manipulation tasks. To fully test the advantages and disadvantages of the this model, it was compared with the MFRL method over various test: utilizing the model in the same environment it was trained; utilizing the model when changes are done on the objects, such as weight and size changes; guiding a pre-trained policy in another task with the correct dynamic model; computational costs of running both methods.\\
As for the main test, performing both models on the environment they were trained in, Model-Based Lookahead achieved a better Average Reward than its policy counterpart for all objects when using the fully-actuated system, but only increase the Average Reward of the egg for the under-actuated system. In the fully-actuated system, not only Average Reward, but also a noticeable change in the Average Consecutive Successes obtained, with the only metric not being affected in all cases being the Average Episode Length, which did not change in the cases where it was already very large, such as the cube and egg. In the under-actuated system, the policy of both the cube and parallelepiped performed very poorly, and utilizing trajectory evaluation did not improve on this. The egg test however did see a large increase in both Average Reward and Average Consecutive Success. The performance of Model-Based Lookahead seems to be dependent on the quality of the policy, as well as the type of system it is being performed on. When the policy fails to achieve good results, the guidance of trajectory evaluation is not going to change the performance significantly, but if the policy already performs well, there is also an increase in Average Reward.\\
As for the generalization test performed on the fully-actuated system, trajectory evaluation seems to increase performance even in large changes in dynamics. It is clear that changing the dynamics of the system, which deviates the calculated trajectories from reality, continuously reduces the gains from applying Model-Based Lookahead. 
When utilizing trajectory evaluation to guide a pre-trained policy, in this case in the cube, to perform a task on both the egg and parallelepiped with their respective dynamic models, the increase in Average Reward was not significant enough to result in any change in Average Consecutive Successes. Similar to results obtained in the under-actuated case, when the policy does not perform well, the gains from Model-Based Lookahead are small.
Finally in terms of computational costs, trajectory evaluation increases the overall runtime, with a large decrease in iterations/second mostly dependent on the horizon value used in the MPC. 
\section*{Acknowledgements}
This work was supported by LARSyS FCT funding (DOI: 10.54499/LA/P/0083/2020, 10.54499/UIDP/50009/2020, and 10.54499/UIDB/50009/2020), the H2020 FET-Open project RePAIR under EU grant agreement 964854, by the Lisbon Ellis Unit.

%

\bibliographystyle{IEEEtran}

\bibliography{references}

\end{document}